\documentclass{article}

    \usepackage[preprint,nonatbib]{neurips_2021}

\usepackage[utf8]{inputenc} 
\usepackage[T1]{fontenc}    
\usepackage[colorlinks]{hyperref}       
\usepackage{url}            
\usepackage{booktabs}       
\usepackage{amsfonts}       
\usepackage{nicefrac}       
\usepackage{microtype}      
\usepackage{xcolor}         

\usepackage{algorithm, algpseudocode}
\usepackage{xspace}
\usepackage{subcaption}
\usepackage{multirow}
\usepackage[pdftex]{graphicx}
\usepackage{enumitem}
\usepackage{amsmath}
\usepackage{bm}
\usepackage{wrapfig}

\newcommand{\BLUE}[1]{\textcolor{blue}{#1}}
\definecolor{fig_red}{RGB}{206,41,71}
\definecolor{fig_blue}{RGB}{74,127,170}
\newcommand{\FIGRED}[1]{\textcolor{fig_red}{#1}}
\newcommand{\FIGBLUE}[1]{\textcolor{fig_blue}{#1}}

\newcommand{\romman}[1]{\uppercase\expandafter{\romannumeral#1}}

\newcommand{\name}[0]{MOAT\xspace}
\newcommand{\pname}[1]{MOAT$_#1$\xspace}
\newcommand{\iname}[0]{IMOAT\xspace}
\newcommand{\piname}[1]{IMOAT$_{#1}$\xspace}

\makeatletter
\DeclareRobustCommand\onedot{\futurelet\@let@token\@onedot}
\def\@onedot{\ifx\@let@token.\else.\null\fi\xspace}

\def\eg{\emph{e.g}\onedot} 
\def\ie{\emph{i.e}\onedot} 
 
\def\etc{\emph{etc}\onedot} 
 
\def\etal{\emph{et al}\onedot}
\makeatother

\title{Multi-stage Optimization based Adversarial Training}

\author{
    Xiaosen Wang\textsuperscript{\rm 1}\thanks{The first two authors contribute equally.}, Chuanbiao Song\textsuperscript{\rm 1}, Liwei Wang\textsuperscript{\rm 2}, Kun He\textsuperscript{\rm 1}\thanks{Corresponding author.}\\
    \textsuperscript{\rm 1} School of Computer Science and Technology,  Huazhong University of Science and Technology \\
    \textsuperscript{\rm 2} School of Electronics Engineering and Computer Sciences, Peking University \\
    \texttt{ \{xiaosen, cbsong\}@hust.edu.cn} \texttt{wanglw@cis.pku.edu.cn} \texttt{brooklet60@hust.edu.cn}
}

\begin{document}

\maketitle

\begin{abstract}

In the field of adversarial robustness, there is a common practice that adopts the single-step adversarial training for quickly developing adversarially robust models. However, the single-step adversarial training is most likely to cause catastrophic overfitting, as after a few training epochs it will be hard to generate strong adversarial examples to continuously boost the adversarial robustness. In this work, 
we aim to avoid the catastrophic overfitting by introducing multi-step adversarial examples during the single-step adversarial training. Then, to balance the large training overhead of generating multi-step adversarial examples, we propose a Multi-stage Optimization based Adversarial Training (MOAT) method that periodically trains the model on mixed benign examples, single-step adversarial examples, and multi-step adversarial examples stage by stage. In this way, the overall training overhead is reduced significantly, meanwhile, the model could avoid catastrophic overfitting. Extensive experiments on CIFAR-10 and CIFAR-100 datasets demonstrate that under similar amount of training overhead, the proposed MOAT exhibits better robustness than either single-step or multi-step adversarial training methods. 
\end{abstract}

\section{Introduction}

Deep neural networks (DNNs) have achieved impressive performance for various machine learning tasks~\cite{he2016deep,girshick2015fast,long2015fully,devlin2018bert}. However, DNNs have been found to be strikingly vulnerable to adversarial examples~\cite{szegedy2013intriguing,goodfellow2014explaining}, in which an imperceptible adversarial perturbation around the benign input data can easily mislead the prediction of DNNs. The threat of adversarial examples has attracted wide attention to the security of deep learning applications, especially on autonomous driving~\cite{eykholt2018robust}, medicine~\cite{finlayson2018adversarial} and face verification~\cite{sharif2016accessorize} \etc.

Many defense methods~\cite{guo2017countering,xie2017mitigating,samangouei2018defense,goodfellow2014explaining,tramer2017ensemble,madry2017towards,zhang2019theoretically} have been proposed to improve the adversarial robustness of DNNs. Among these methods, Projected Gradient Descent Adversarial Training (PGD-AT)~\cite{madry2017towards} has been demonstrated as one of the most effective  methods~\cite{athalye2018obfuscated,dong2020benchmarking}. To train a robust model, PGD-AT alternatively generates adversarial examples with multi-step projected gradient descent and optimizes the model parameters by training on the generated adversarial examples. However, a significant drawback lies that PGD-AT is highly time-consuming, mainly because the multi-step attack significantly increases the training overhead.

A straightforward remedy for this issue is to implement adversarial training with the single-step attacks (\eg Fast Gradient Sign Method (FGSM)) to reduce the training overhead~\cite{goodfellow2014explaining,wong2020fast,andriushchenko2020understanding,kim2021understanding}. Nevertheless, single-step adversarial training could lead to \textit{catastrophic overfitting}~\cite{wong2020fast,andriushchenko2020understanding}, in which the robust accuracy of the model against strong iterative attack (\eg PGD) would suddenly decrease to almost zero after a few training epochs. While a few methods~\cite{kim2021understanding,andriushchenko2020understanding} have been proposed to avoid catastrophic overfitting, these methods would increase the training overhead or hurt the standard accuracy compared with single-step adversarial training.

\begin{figure}[tb]
    \centering
    \begin{minipage}[c]{0.45\textwidth}
        \begin{subfigure}{\textwidth}
            \includegraphics[width=\linewidth]{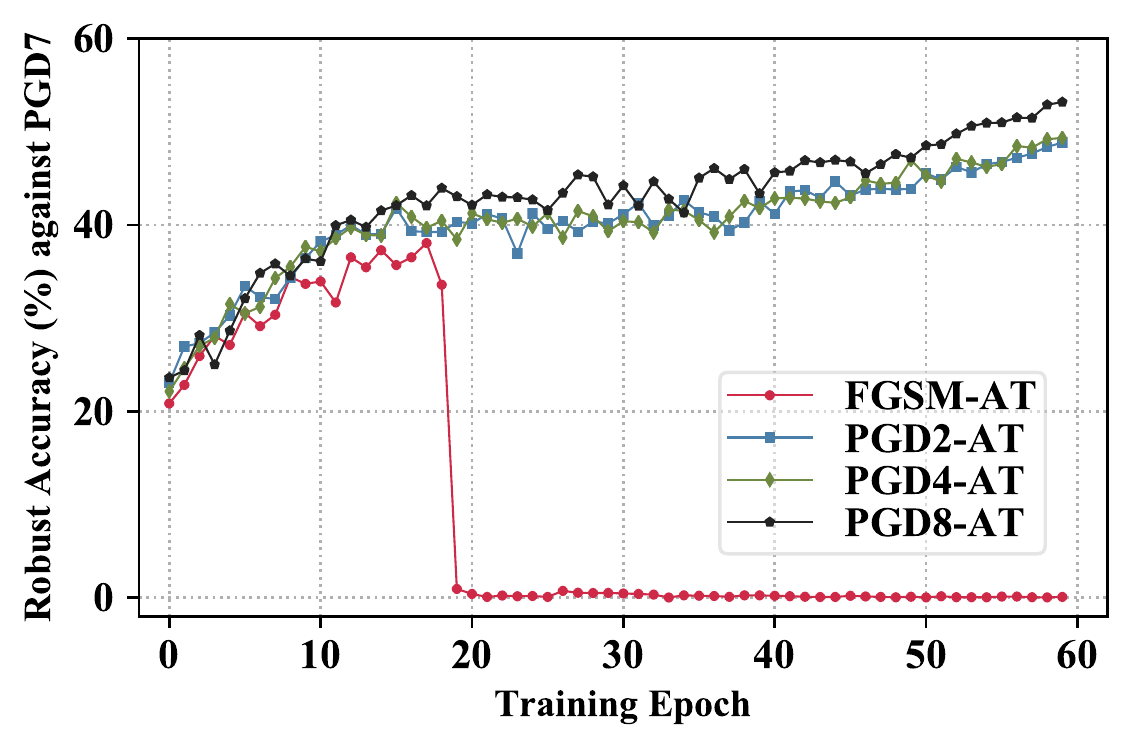}
            \caption{CIFAR-10}
        \end{subfigure}
    \end{minipage}
    \hspace{0.3cm}
    \begin{minipage}[c]{0.45\textwidth}
        \begin{subfigure}{\textwidth}
            \includegraphics[width=\linewidth]{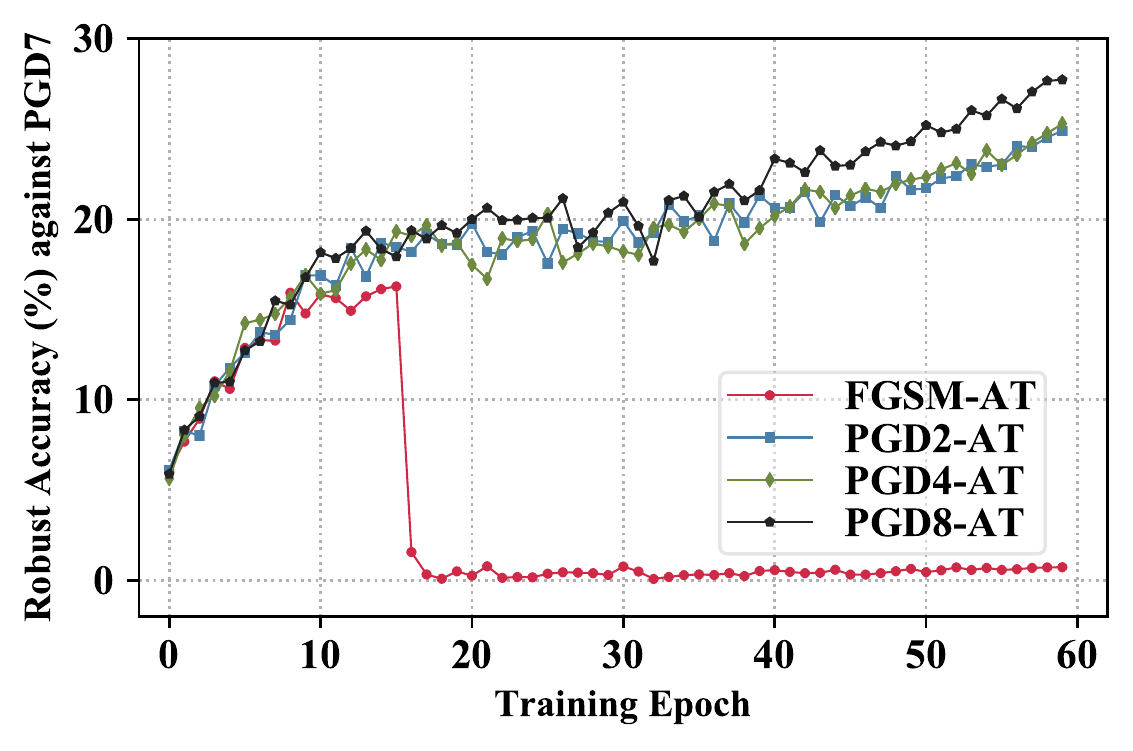}
            \caption{CIFAR-100}
        \end{subfigure}
    \end{minipage}
    \caption{Comparisons of the robust test accuracy against PGD7 attack for FGSM-AT, PGD2-AT, PGD4-AT and PGD8-AT on CIFAR-10 and CIFAR-100 datasets for each training epoch. The detailed experimental settings are summarized in Section~\ref{sec:Connection}. The robust test accuracy of single-step adversarial training FGSM-AT (\FIGRED{red line}) decreases rapidly as the training progresses. By contrast, using strong adversarial examples generated by multi-step attacks (\eg PGD2 attack, \FIGBLUE{blue line}) prevents catastrophic overfitting, achieves and maintains the high robust test accuracy.
}
    \label{fig:observation}
\end{figure}

In this work, we analyze the relationship between catastrophic overfitting and adversarial examples used in training. 
A key observation is that adopting multi-step adversarial examples in training can prevent models from catastrophic overfitting, even using the adversarial examples generated by the two-step PGD attack. 
For instance, as shown in Figure~\ref{fig:observation} (a), on the CIFAR-10 dataset, for the single-step adversarial training (\ie FGSM-AT), the adversarial test accuracy against PGD7 attack suddenly decreases to 
almost zero at the 19-th training epoch. By contrast, the multi-step adversarial training methods (\eg PGD2-AT) do not
suffer from the catastrophic overfitting and exhibit increasing robustness throughout the training process. We see similar trends on the CIFAR-100 dataset, as shown in Figure~\ref{fig:observation} (b).


Motivated by this observation, we consider avoiding the catastrophic overfitting by using multi-step adversarial examples during the single-step adversarial training. 
However, generating multi-step adversarial examples would introduce higher computational overhead.
To reduce the overall training overhead and prevent the models from catastrophic overfitting, we propose a Multi-stage Optimization based Adversarial Training (\name) method by enabling the use of benign examples, single-step adversarial examples, and multi-step adversarial examples. Recalling that multi-stage optimization has been investigated in curriculum learning~\cite{bengio2009curriculum} and transfer learning~\cite{bengio2012deep}, where the training process is divided into several sub-tasks to be completed stage by stage.
However, there are few explorations in adversarial training, among which, Hendrycks \etal~\cite{hendrycks2019using} try to adopt pre-trained model parameters as the initialization of the current model parameters to improve the adversarial robustness.

In this paper, we aim to explore the multi-stage optimization in adversarial training for quickly developing adversarially robust models.
Unlike traditional adversarial training methods that utilize a fixed type of adversarial examples, our \name periodically trains the model on benign examples, single-step adversarial examples and multi-step adversarial examples stage by stage. 
In this way, the solution found on each type of training samples could serve as a prior to the next training stage, which could improve the generalization on the different types of samples and reduce the overall training overhead. 
Extensive experiments on CIFAR-10 and CIFAR-100 datasets validate the efficiency and effectiveness of \name on developing adversarially robust models, and show that \name could significantly narrow the performance gap between single-step and multi-step adversarial training methods.






\section{Related Work}
Given a classfication dataset $\mathcal{D}=\{(x_i, y_i)\}_{i=1}^N$ with $N$ training samples, the goal of adversarial attack is to find an adversarial example $x_{adv}\in \mathcal{B}_\epsilon(x)=\{x' : \|x'-x\|_p\leq \epsilon\}$ for each sample $x$ to mislead the target model $f_\theta$ with parameters $\theta$, where $\epsilon$ denotes the magnitude of  perturbation, $p$ denotes the norm and we focus on $p=\infty$ in this work. 
Since Szegedy~\etal~\cite{szegedy2013intriguing} identified the vulnerability of DNNs to adversarial example, various adversarial attacks have been proposed, such as gradient-based attack~\cite{goodfellow2014explaining,kim2021understanding,madry2017towards,carlini2017towards,croce2020reliable}, transfer-based attack~\cite{liu2016delving,dong2018boosting,xie2019improving,wang2021enhancing}, score-based attack~\cite{ilyas2018black,ilyas2018prior,cheng2019improving,yan2019subspace} and decision-based attack~\cite{brendel2017decision,cheng2018query,li2020qeba,chen2020boosting}. 

On the other hand, to mitigate the threat of adversarial attacks, a variety of defense methods have been proposed, \eg adversarial training~\cite{goodfellow2014explaining,tramer2017ensemble,madry2017towards,zhang2019theoretically,song2019improving,wang2019improving,ding2018mma}, input pre-processing~\cite{guo2017countering,xie2017mitigating,samangouei2018defense,song2017pixeldefend}, certified defense~\cite{raghunathan2018certified,raghunathan2018semidefinite,cohen2019certified,gowal2019scalable} \etc. Among which, adversarial training, which improves adversarial robustness by training on adversarial examples, has been shown as one of the most effective methods~\cite{athalye2018obfuscated,croce2020reliable}.
Adversarial training can be formulated as a minimax optimization problem~\cite{madry2017towards}:
\begin{equation}
\label{eq:minmax}
    \min_{\theta} \mathbb{E}_{(x,y)\sim \mathcal{D}} [\max_{x_{adv} \in \mathcal{B}(x)} \mathcal{L}(f_\theta(x_{adv}), y)],
\end{equation}
where $\mathcal{L}(f_\theta(x),y)$ is the loss function (\eg cross entropy loss) of the target model $f$. 

Goodfellow~\etal~\cite{goodfellow2014explaining} first propose a single-step adversarial training, Fast Gradient Sign Method Adversarial Training (FGSM-AT) by solving the inner maximization problem of Eq.~\ref{eq:minmax} as follows:
\begin{equation}
    x_{adv} = x + \epsilon \cdot {\rm sign}(\nabla_{x} \mathcal{L}(f_\theta(x), y)).
\end{equation}
For a long time after that, it is commonly believed that FGSM-AT cannot provide real robustness against iterative adversarial attacks~\cite{madry2017towards}. 
To address this issue, Mardy~\etal~\cite{madry2017towards} introduce an iterative variant of FGSM-AT, namely Projected Gradient Descent Adversarial Training (PGD-AT). For PGD-AT, the solution for the inner maximization problem of Eq.~\ref{eq:minmax} is formulated as follows:
\begin{equation}
    x_{adv}^{t+1} = \Pi_{\mathcal{B}_\epsilon(x)} [x_{adv}^t + \alpha \cdot {\rm sign}(\nabla_x \mathcal{L}(f_\theta(x_{adv}^t), y)],
\end{equation}
where $\alpha$ is the step size for each attack iteration, $\Pi_{\mathcal{B}_\epsilon(x)}(\cdot)$ denotes the projection function, and $x_{adv}^0 = x + \eta, \ \eta \sim \mathcal{U}(-\epsilon, \epsilon)$ is the initial perturbed example. 
Extensive works have shown that PGD adversarial training (PGD-AT) leads to robust models against powerful adversarial attacks~\cite{athalye2018obfuscated,croce2020reliable,dong2020benchmarking}. 


Nevertheless, PGD-AT also introduces a high training overhead, which is much higher than the standard training and FGSM-AT due to the iterative solution for the inner maximization problem. 
To accelerate the adversarial training, Shafahi~\etal~\cite{shafahi2019adversarial} propose free adversarial training that utilizes single back-propagation to update both the model parameters and adversarial perturbations simultaneously. Zhang \etal \cite{zhang2019you} restrict most of the forward and backward propagation within the first layer of the neural networks during the adversary generation so as to reduce the training overhead. 

Recently, Wong~\etal~\cite{wong2020fast} identify \textit{catastrophic overfitting} in the single-step adversarial training FGSM-AT, in which the robust accuracy against iterative attack (\eg PGD7) suddenly drops to almost $0\%$ after a certain number of training epochs. To mitigate this issue, they propose Fast Adversarial Training (Fast-AT) which adopts early stopping to avoid catastrophic overfitting, and incorporates a larger step size $\alpha$ and random start $\eta\sim\mathcal{U}(-\epsilon, \epsilon)$ into FGSM to solve the inner maximization problem:
\begin{equation}
    x_{adv} = \Pi_{\mathcal{B}_\epsilon(x)} [x+\eta +\alpha \cdot {\rm sign}(\nabla_x \mathcal{L}(f_\theta(x+\eta),y))].
\end{equation}

Later on, a line of work focuses on avoiding catastrophic overfitting in Fast-AT. Andriushchenko~\etal~\cite{andriushchenko2020understanding} argue that the random start in Fast-AT only improves the linear approximation quality and propose GradAlign to prevent catastrophic overfitting for FGSM-AT. Vivek~\etal~\cite{vivek2020single} 
propose dropout scheduling for single-step adversarial training to achieve reasonable robustness. Li~\etal~\cite{li2020towards} adopt the PGD-10 attack to help Fast-AT recover from catastrophic overfitting. Kim~\etal~\cite{kim2021understanding} find that catastrophic overfitting is closely related to decision boundary distortion and propose to utilize the minimal scaled adversarial perturbation for training to prevent catastrophic overfitting.


\section{Methodology}
In this section, we first analyze the relationship between catastrophic overfitting and adversarial examples used in the training. Based on the analysis, we introduce the proposed Multi-stage Optimization based Adversarial Training (\name) method to quickly develop adversarially robust models by enabling the use of benign examples, single-step adversarial examples, and multi-step adversarial examples, and provide an improved variant of \name, denoted as \iname.

\subsection{Relationship between Catastrophic Overfitting and Adversarial Examples}
\label{sec:Connection}

Previous works~\cite{wong2020fast,andriushchenko2020understanding} have shown that using single-step adversarial examples in adversarial training leads to catastrophic overfitting, which is a critical challenge for computationally efficient adversarial training. However, PGD-AT which adopts multi-step adversarial examples in adversarial training, does not have such a problem. To gain more insights into the phenomenon of catastrophic overfitting in adversarial training, we aim to analyze the relationship between catastrophic overfitting and the adversarial strength of adversarial examples used in the training and try to answer the following question:

\textit{
How to choose a large enough value for $K$ to avoid catastrophic overfitting when training on the $K$-step adversarial examples ($K \geq  1$)?}

To this end, we record the robust test accuracy of single-step adversarial training (\ie FGSM-AT) and $K$-step adversarial training (\ie, PGDK-AT) on CIFAR-10 and CIFAR-100 datasets. 
The maximum perturbation $\epsilon$ used in training is fixed to $8/255$ and the step size $\alpha$ for PGDK-AT is fixed to $\max(\epsilon/4, \epsilon/K)$.
To verify the robust test accuracy, we use PGD7 attack with the same maximum perturbation $\epsilon=8/255$ and step size $\alpha = 2/255$. 

As shown in Figure~\ref{fig:observation}, we illustrate the robust test accuracy for various adversarial training methods (\ie, FGSM-AT, PGD2-AT, PGD4-AT and PGD8-AT) over each training epoch. 
We can clearly observe the catastrophic overfitting in single-step adversarial training FGSM-AT, in which the robust test accuracy against the PGD7 attack suddenly decreases to almost zero after 19 training epochs on CIFAR-10 dataset, and a similar phenomenon also happens on CIFAR-100 dataset.
On the other hand, we notice that there is no catastrophic overfitting for any $K$-step adversarial training ($K \geq 2$) through the entire training process and the larger value for $K$ can lead to the higher robust accuracy. Hence, it is clear that adopting stronger (multi-step) adversarial examples could effectively avoid the catastrophic overfitting compared with single-step adversarial examples.

\subsection{Multi-stage Optimization based Adversarial Training}
\label{sec:method}

Motivated by the above observation, we consider adopting multi-step adversarial examples to avoid catastrophic overfitting in single-step adversarial training. 
However, note that the training overhead of adversarial training is linear to the number of back-propagation~\cite{madry2017towards, wong2020fast}, 
when adopting $K$-step adversarial examples during training, the total number of back-propagation is proportional to $(K+1)T$, where $T$ is the number of training epochs. 
As such, the single-step adversarial training FGSM-AT's training overhead is $2T$, while the multi-step adversarial training PGD2-AT's is $3T$ which is roughly $1.5\times$ slowdown than FGSM-AT. Thus, introducing multi-step adversarial training samples would increase the training overhead of the single-step adversarial training.

To balance the high training overhead of introducing multi-step adversarial examples, we propose a Multi-stage Optimization based Adversarial Training (\name) method that periodically trains the model on benign examples, single-step adversarial examples, and multi-step adversarial examples stage by stage.
The intuition behind \name is that the solution found at each stage can serve as a prior to the next stage, (\ie, the model parameters optimized in each stage can serve as the pre-trained model parameters for the next stage), which could improve the generalization of the model and reduce the overall training overhead.

Specifically, at the initial stage (Stage \romman{1}) of \name, the learning on benign examples aims to improve the standard generalization of the model, 
which does not generate adversarial examples and takes less training overhead. 
Based on the model with great standard generalization, at Stage \romman{2} of \name, the learning on single-step adversarial examples aims to train a robust model with few back-propagations.
At the third stage (Stage \romman{3}) of \name, 
the learning on multi-step adversarial examples is used to prevent the model from catastrophic overfitting with more back-propagations. In this way, \name would reduce the overall training overhead while avoiding catastrophic overfitting through the entire training process. 
We periodically perform the proposed multi-stage optimization over these stages to prevent models from overfitting to a certain type of training samples to obtain a more robust model with less training overhead.

The implementation details of \name are summarized in Algorithm~\ref{alg:MSOAT}.
For the Stage \romman{1} of \name, we conjecture
that \textit{the higher generalization obtained at Stage \romman{1} would help the following stages find stronger adversarial examples to train a more robust model instead of using the misclassified clean examples for training}. Thus, we use \textit{mixup} examples~\cite{zhang2017mixup} as the benign examples at Stage \romman{1} to promote the standard generalization of the model.
Since \textit{mixup} just pre-processes the training samples by interpolation, the computational overhead could be negligible regarding to the overall training overhead.
Note that the training process of \name would end at Stage \romman{3} of the last round to obtain the adversarially robust model.

    

\begin{algorithm}[tb]
    \algnewcommand\algorithmicinput{\textbf{Input:}}
    \algnewcommand\Input{\item[\algorithmicinput]}
    \algnewcommand\algorithmicoutput{\textbf{Output:}}
    \algnewcommand\Output{\item[\algorithmicoutput]}
    \caption{Multi-stage Optimization based Adversarial Training (\name) }
    \label{alg:MSOAT}
	\begin{algorithmic}[1]
		\Input $\mathcal{D}$: training datasets, $T$: training epochs, $b$: training batch size, 
		$\eta$: learning rate,
		$K$: number of iteration steps for multi-step adversarial examples,
		$\alpha_m$: step size of the multi-step adversarial examples,
		$\alpha_s$: step size of the single-step adversarial examples,
		$\epsilon$: maximum perturbation.
		\Output Adversarially robust model $f_\theta$
		\State Randomly Initialize $\theta = \theta_0$
		\For{$t = 0 \ \ {\rm to}\ \  T-1$}
		    \State  Stage $s = t\ \%\ 3 + 1$
		    \For{each mini-batch $(x_{b}, y_{b}) \in \mathcal{D}$}
	        \If{$s == 1$}  \algorithmiccomment{Stage \romman{1}}
	        \State Obtain $(x'_{b}, y'_{b})$ by randomly shuffling $(x_{b}, y_{b})$, and sample $\lambda\sim U(0,1)$
	        \State Mixup $(x_{b}, y_{b})$ with $(x'_{b}, y'_{b})$:
	        \State \qquad $\hat{x}_{b}  = \lambda \cdot x_{b} + (1-\lambda) \cdot x'_{b}$, \quad $\hat{y}_{b} = \lambda \cdot y_{b} + (1-\lambda) \cdot y'_{b}$ 
	        \ElsIf{$s == 2$} \algorithmiccomment{Stage \romman{2}}
	        \State Generate the single-step adversarial training samples:
	        \State \qquad $x_b^0 = \Pi_{\mathcal{B}_\epsilon(x_b)}[x_b + \mathcal{U}(-\epsilon, \epsilon)]$
	        \State \qquad $\hat{x}_b = \Pi_{\mathcal{B}_\epsilon(x_b)}[x_b^0+\alpha_s \cdot {\rm sign}(\nabla_{x_b} \mathcal{L}(f_\theta(x^0_{b}), y_b)]$, \quad $\hat{y}_{b} = y_{b}$
	        \ElsIf{$s == 3$} \algorithmiccomment{Stage \romman{3}}
	        \State Generate the multi-step adversarial training samples:
	        \State \qquad $x_b^0 = \Pi_{\mathcal{B}_\epsilon(x_b)}[x_b + \mathcal{U}(-\epsilon, \epsilon)]$, \quad $\alpha = \max(\alpha_m, \epsilon/K) $
	        \State \qquad \textbf{for}\ \ $k = 1$ \ \ {\rm to}\ \ $K$  \textbf{do}
	        \State \qquad\qquad $x^k_{b}=\Pi_{\mathcal{B}_\epsilon(x_b)}[x_b^{k-1}+\alpha \cdot {\rm sign}(\nabla_{x_b} \mathcal{L}(f_\theta(x^{k-1}_{b}), y_b)]$
            \State \qquad \textbf{end for}
	        \State \qquad $\hat{x}_b = x^K_b$, \quad $\hat{y}_{b} = y_{b}$
	        \EndIf
	        \State $\theta_{t+1} \leftarrow \theta_{t} - \eta \cdot  \nabla_{\theta_t} \mathcal{L}(f_{\theta_t}(\hat{x}_{b}), \hat{y}_{b})$  \algorithmiccomment{Update the model parameters}
	        \EndFor
		\EndFor
	
	\end{algorithmic} 
\end{algorithm}

For convenience, we denote the proposed method \name using $K$-step adversarial examples at Stage \romman{3} as \pname{K}. Intuitively, increasing $K$ would promote the adversarial robustness but also increase the overall training overhead, which we will discuss in Section~\ref{sec:ablation}. Specifically, the average training overhead of \pname{K} is $[1+2+(K+1)]T/3$. 
As such, the overall training overhead  (\ie $2T$) of \pname{2} is roughly equal to that of the single-step adversarial training FGSM-AT, which we will show in Section~\ref{sec:stepwise} to ~\ref{sec:cycle}.

\subsection{Improved Multi-stage Optimization based Adversarial Training}
\label{sec:imoat}
Furthermore, we conjecture that \textit{as the training progresses, the model exhibits better robustness and needs more powerful adversarial examples to further boost the robustness}. Based on this hypothesis, we further propose an improved variant of \name, \iname, with an increasing step $K$ for the multi-step adversarial examples. Specifically, we split the overall training process into three equal training phases. In the early training phase, we adopt a small number of steps $K_1$. Then we adopt a medium number of steps $K_2$ in the middle training phase. Finally, we adopt a large number of steps $K_3$ in the later training phase. 
For simplicity, we denote the \iname with hyper-parameters ($K_1,K_2,K_3$) as \piname{{K_1,K_2,K_3}}. The average training overhead of \piname{{K_1,K_2,K_3}} is  $\{1+2+[(K_1+1)+(K_2+1)+(K_3+1)]/3\}T/3$.
As such, the average training overhead (\ie $2T$) of \piname{{1,2,3}} is almost equal to that of the single-step adversarial training FGSM-AT.
The implementation details of the \iname are summarized in Algorithm~\ref{alg:MSOAT++} in Appendix~\ref{lab:msoat++}.




\section{Experiments}
We conduct extensive empirical evaluations to validate the effectiveness and efficiency of the proposed methods, including the comparisons with other adversarial training methods using step-wise or cyclic learning rate schedules and the ablation studies. 

\subsection{Experimental Setup}
\textbf{Datasets and Models.} We conduct experiments on two widely used benchmark datasets, namely CIFAR-10~\cite{krizhevsky2009learning} and CIFAR-100~\cite{krizhevsky2009learning}. CIFAR-10 consists 10 classes with 5,000 $32\times 32$ training images and 1,000 testing images per class. CIFAR-100 contains 100 classes where each class has 500 training images and 100 test images. To align with previous adversarial training methods~\cite{wong2020fast,andriushchenko2020understanding,kim2021understanding}, we adopt PreAct ResNet-18~\cite{he2016deep} as the neural network.

\textbf{Baselines.} To verify the effectiveness of our methods (\ie \name and \iname), we compare the methods with various single-step adversarial training methods, namely FGSM-AT~\cite{goodfellow2014explaining}, Fast-AT~\cite{wong2020fast}, GradAlign~\cite{andriushchenko2020understanding} and Kim~\etal~\cite{kim2021understanding} and multi-step adversarial training PGD2-AT~\cite{madry2017towards}, PGD7-AT~\cite{madry2017towards}.

\textbf{Hyper-parameters.}  We set the regularization hyper-parameter to $0.2$ for GradAlign~\cite{andriushchenko2020understanding}, set the number of checkpoints to $3$ for Kim~\etal~\cite{kim2021understanding}, and set the step size to $1.25\epsilon$ for Fast-AT~\cite{wong2020fast} and Kim~\etal~\cite{kim2021understanding}.
Following previous works~\cite{wong2020fast,kim2021understanding,madry2017towards}, for our method \name, we use $\alpha_s = 1.25\epsilon$ for single-step adversarial examples, and $\alpha_m = \epsilon/4$ for multi-step adversarial examples.

\textbf{Training Details.} We follow the training settings in \cite{kim2021understanding}, in which we use SGD with momentum of 0.9 and weight decay of $5\times10^{-4}$ as the optimizer and batch size of 128. 
To thoroughly validate the effectiveness, we adopt two popular learning rate schedules used in adversarial training, namely step-wise learning rate and cyclic learning rate, and the maximum learning rate is $0.2$. 
For the step-wise learning rate, we set the total epoch to 200, and the learning rate decays with a factor of 0.1 at 60, 120, and 160 epochs. For the cyclic learning rate, the total epoch is set to 60, and the learning rate linearly increases at half of the total epochs. The maximum magnitude $\epsilon$ of the perturbation used in training is $8/255$.
All the experiments are performed on a single NVIDIA Tesla T4 GPU.

\textbf{Evaluation Metrics.} We consider both standard accuracy and robust accuracy for the evaluation. For the robust accuracy, we evaluate the models under various white-box attacks, namely FGSM~\cite{goodfellow2014explaining}, PGD~\cite{madry2017towards}, Carlini-Wagner (CW)~\cite{carlini2017towards}, Momentum Iterative Method (MIM)~\cite{dong2018boosting} and AutoAttack (AA)~\cite{croce2020reliable}. The maximum iteration for PGD, CW, and MIM is set to 50. Both PGD and CW adopt 10 random restarts. 
The CW attack adopts the PGD optimization with the margin-based loss function~\cite{carlini2017towards}. The maximum magnitude of the perturbation is set to $8/255$, which is the same as in the training.

\begin{table}[tb]
  \caption{Classification accuracy (\%) and training time (min) of multi-step, single-step adversarial training methods and the proposed methods (\ie \name and \iname) using step-wise learning rates on CIFAR-10 and CIFAR-100 datasets against white-box attacks with $\epsilon=8/255$. We highlight the highest classification accuracy among each type of adversarial training in \textbf{bold} and the results of FGSM-AT and Fast-AT in \BLUE{blue} to emphasize the occurrence of catastrophic overfitting.}
  \label{tab:stepwise}
  \centering
    \begin{subtable}{\textwidth}
        \caption{Evaluations on CIFAR-10.}
        \label{tab:stepwise:CIFAR10}
        \resizebox{\textwidth}{!}{
            \begin{tabular}{c|cccccccc}
                \toprule
                & Method & Clean & FGSM &PGD50 & CW50 & MIM50 & AA & Time(min)\\
                \midrule
                
                \multirow{4}{*}{Single-step} 
                & FGSM-AT & \textbf{\BLUE{90.87}} & \BLUE{83.06} & ~~\BLUE{0.55} & ~\BLUE{0.62} & ~~\BLUE{3.77} & ~~\BLUE{0.01} & ~~411.87\\
                & Fast-AT & \BLUE{86.12} & \textbf{\BLUE{93.55}} & ~~\BLUE{0.07} & ~~\BLUE{0.13} & ~~\BLUE{2.29} & ~~\BLUE{0.00} & ~~402.13 \\
                & GradAlign & 82.78 & 45.06 & \textbf{33.37} & \textbf{35.30} & 41.30 & \textbf{31.63} & ~~973.97\\
                & Kim~\etal & 87.37 & 47.63 & 31.91 & 32.33 & \textbf{41.61} & 30.51 & ~~438.37\\
                \midrule 
                
                \multirow{2}{*}{Multi-step} 
                & PGD2-AT & \textbf{85.96} & 47.20 & 34.83 & 35.56 & 42.70 & 33.61 & ~~595.53\\
                & PGD7-AT & 83.35 & \textbf{49.62} & \textbf{39.01} & \textbf{39.87} & \textbf{46.11} & \textbf{37.89} & 1565.33\\
                \midrule
                
                \multirow{4}{*}{Multi-stage} 
                & \pname{2} & 87.61 & 50.25 & 34.35 & 36.12 & 44.83 & 32.57 & ~~410.09\\
                & \piname{1,2,3} & \textbf{87.63} & 50.02 & 34.36 & 36.20 & 44.35 & 32.72 & ~~413.34\\
                & \pname{5} & 86.81 & 51.54 & 36.92 & 38.81 & 46.68 & 35.53 & ~~603.03\\
                & \piname{2,5,8} & 86.76 & \textbf{51.99} & \textbf{38.28} & \textbf{39.61} & \textbf{46.85} & \textbf{36.19} & ~~619.42\\
                \bottomrule
            \end{tabular}
        }
    \end{subtable}
    \begin{subtable}{\textwidth}
        \caption{Evaluations on CIFAR-100.}
        \label{tab:stepwise:CIFAR100}
        \resizebox{\textwidth}{!}{
            \begin{tabular}{c|cccccccc}
                \toprule
                & Method & Clean & FGSM &PGD50 & CW50 & MIM50 & AA & Time(min)\\
                \midrule
                
                \multirow{4}{*}{Single-step} 
                & FGSM-AT & \textbf{\BLUE{66.15}} & \BLUE{45.50} & ~~\BLUE{0.84} & ~~\BLUE{0.97} & ~~\BLUE{1.63} & ~~\BLUE{0.05} & ~~408.30\\
                & Fast-AT & \BLUE{57.92} & \textbf{\BLUE{86.20}} & ~~\BLUE{0.00} & ~~\BLUE{0.00} & ~~\BLUE{0.01} & ~~\BLUE{0.00} & ~~407.87\\
                & GradAlign & 55.40 & 20.02 & 13.81 & \textbf{15.16} & 17.95 & \textbf{12.85} & 1053.03\\
                & Kim~\etal & 62.34 & 22.32 & \textbf{13.82} & 14.71 & \textbf{19.29} & 12.79 & ~~421.87\\
                \midrule 
                
                \multirow{2}{*}{Multi-step} 
                & PGD2-AT & \textbf{57.97} & 21.51 & 15.23 & 16.13 & 19.35 & 14.42 & ~~596.50\\
                & PGD7-AT & 55.66 & \textbf{23.43} & \textbf{17.69} & \textbf{18.70} & \textbf{21.64} & \textbf{17.04} & 1551.03\\
                \midrule
                
                \multirow{3}{*}{Multi-stage}
                & \pname{2} & 61.96 & 24.35 & 16.02 & 16.76 & 21.43 & 14.44 & ~~410.19\\
                & \piname{1,2,3} & \textbf{62.11} & 24.28 & 15.77 & 16.47 & 21.46 & 14.06 & ~~414.38\\
                & \pname{5} & 61.19 & \textbf{24.65} & 16.93 & 17.41 & 22.12 & 15.07 & ~~608.49\\
                & \piname{2,5,8} & 61.51 & 24.53 & \textbf{17.28} & \textbf{17.99} & \textbf{22.16} & \textbf{15.72} & ~~614.74\\
                \bottomrule
            \end{tabular}
        }
    \end{subtable}
\end{table} 
\subsection{Evaluations on Step-wise Learning Rate Schedule}
\label{sec:stepwise}

The step-wise learning rate schedule has been widely used in adversarial training~\cite{madry2017towards,tramer2017ensemble,zhang2019theoretically}. We first evaluate the robustness of the proposed methods and the baselines against various white-box attacks using the step-wise learning rate schedule. we report the standard accuracy on the clean test data, and the robust test accuracy against various attacks for the proposed methods (\ie \pname{2}, \piname{1,2,3}, \pname{5}, and \piname{2,5,8}) and the baseline methods in Table~\ref{tab:stepwise:CIFAR10}.

On the CIFAR-10 dataset, as shown in Table~\ref{tab:stepwise:CIFAR10}, we can see that: 
\begin{itemize}[leftmargin=*]
\item For the single-step adversarial training methods, FGSM-AT and Fast-AT, the robust accuracy against FGSM attack is very high. However, they exhibit little robustness against other strong iterative attacks due to the catastrophic overfitting. While GradAlign yields higher robust accuracy, it hurts the standard accuracy and significantly increases the training time cost. 
With similar training time cost of Fast-AT, our method \pname{2} achieves higher robust accuracy than the best single-step adversarial training method, \ie GradAlign. 
It is noted that GradAlign needs more than $2\times$ training time cost as compared with our method \pname{2}, due to the double back-propagation introduced by the gradient regularization. 
In addition, our \piname{1,2,3} could achieve slightly better robustness than \pname{2} without extra cost, which supports our hypothesis that we need more powerful adversarial attacks to craft adversarial examples for training in the later training epochs to achieve better robustness.

\item For the multi-step adversarial training methods, PGD7-AT achieves better robustness than PGD2-AT because of the more optimization iterations for adversary generation, leading to nearly $3\times$ training time cost than PGD2-AT. 
With the similar training cost as PGD2-AT, our \pname{5} achieves higher clean and robust accuracy than PGD2-AT, and \piname{2,5,8} could further improve the performance without much extra training overhead. 
More importantly, our \piname{2,5,8} could achieve similar robustness to PGD7-AT (PGD7-AT needs $2.5\times$ training time compared with \piname{2,5,8}), which further validates the  superiority of our methods.
\end{itemize}




    
    

As shown in Table \ref{tab:stepwise:CIFAR100}, on CIFAR-100 dataset, the evaluation results exhibit the same trends as on CIFAR-10 and consistently demonstrate that \name can outperform  other single-step adversarial training methods with similar (or less) training time cost.

Overall, compared with single-step adversarial training and multi-step adversarial training (\ie PGD2-AT), our methods can achieve much better robustness with similar (or less) training time cost. In addition, on the two benchmark datasets, our \piname{2,5,8} exhibits much better performance than GradAlign, and narrow the performance gap between single-step and multi-step adversarial training using much less training time. Besides, the performance improvement of \iname on \name also indicates a new avenue for balancing the robust accuracy and training time cost.


\subsection{Evaluations on Cyclic Learning Rate Schedule}
\label{sec:cycle}
\begin{table}[t]
  \caption{Classification accuracy (\%) and training time (min) of multi-step, single-step adversarial training methods and the proposed methods (\ie \name and \iname) using cyclic learning rates on CIFAR-10 and CIFAR-100 datasets against white-box attacks with $\epsilon=8/255$. We highlight the highest classification accuracy among each type of adversarial training in \textbf{bold} and the results of FGSM-AT, Fast-AT and Kim~\etal in \BLUE{blue} to emphasize the occurrence of catastrophic overfitting.}
  \label{tab:cycle}
  \centering
    \begin{subtable}{\textwidth}
        \caption{Evaluations on CIFAR-10.}
        \label{tab:cycle:CIFAR10}
        \resizebox{\textwidth}{!}{
            \begin{tabular}{c|cccccccc}
                \toprule
                & Method & Clean & FGSM &PGD50 & CW50 & MIM50 & AA & Time(min)\\
                
                \midrule
                \multirow{4}{*}{Single-step} 
                & FGSM-AT & \textbf{\BLUE{93.22}} & \BLUE{31.67} & ~~\BLUE{0.00} & ~~\BLUE{0.00} & ~~\BLUE{0.00} & ~~\BLUE{0.00} & 120.86\\
                & Fast-AT & \BLUE{78.58} & \BLUE{\textbf{99.62}} & ~~\BLUE{0.00} & ~~\BLUE{0.00} & ~~\BLUE{0.00} & ~~\BLUE{0.00} & 123.07\\
                & GradAlign & 85.63 & 55.15 & \textbf{44.73} & \textbf{45.61} & \textbf{51.57} & \textbf{42.25} & 292.76\\
                & Kim~\etal & 90.10 & 53.21 & 38.39 & 39.44 & 48.00 & 36.95 & 132.23\\
                \midrule 
                
                \multirow{2}{*}{Multi-step} 
                & PGD2-AT & \textbf{87.72} & 55.75 & 42.98 & 44.85 & 51.81 & 41.87 & 179.06\\
                & PGD7-AT & 84.76 & \textbf{58.01} & \textbf{48.95} & \textbf{49.26} & \textbf{55.22} & \textbf{46.88} & 464.30\\
                
                \midrule
                \multirow{4}{*}{Multi-stage} 
                & \pname{2} & 87.33 & 54.91 & 43.46 & 44.39 & 51.42 & 41.36 & 123.27\\
                & \piname{1,2,3} & \textbf{87.38} & 55.12 & 44.47 & 45.48 & 51.53 & 42.37 & 123.12\\
                & \pname{5} & 86.23 & 55.91 & 46.12 & 46.91 & 52.76 & 43.71 & 182.15\\
                & \piname{2,5,8} & 84.79 & \textbf{57.00} & \textbf{48.97} & \textbf{48.60} & \textbf{54.40} & \textbf{45.68} & 183.31\\
                \bottomrule
            \end{tabular}
        }
    \end{subtable}
    \begin{subtable}{\textwidth}
        \caption{Evaluations on CIFAR-100.}
        \label{tab:cycle:CIFAR100}
        \resizebox{\textwidth}{!}{
            \begin{tabular}{c|cccccccc}
                \toprule
                & Method & Clean & FGSM &PGD50 & CW50 & MIM50 & AA & Time(min)\\
                \midrule
                \multirow{4}{*}{Single-step} 
                & FGSM-AT & \BLUE{70.11} & \BLUE{26.36} & ~~\BLUE{0.11} & ~~\BLUE{0.02} & ~~\BLUE{0.74} & ~~\BLUE{0.00} & 120.49\\
                & Fast-AT & \BLUE{60.14} & \BLUE{78.77} & ~~\BLUE{0.03} & ~~\BLUE{0.00} & ~~\BLUE{0.53} & ~~\BLUE{0.00} & 123.59\\
                & GradAlign & 59.91 & 29.03 & \textbf{23.78} & \textbf{23.69} & \textbf{27.63} & \textbf{20.74} & 315.40\\
                & Kim~\etal & \BLUE{\textbf{73.29}} & \BLUE{\textbf{78.93}} & ~~\BLUE{0.03} & ~~\BLUE{0.00} & ~~\BLUE{0.52} & ~~\BLUE{0.00} & 126.84\\
                \midrule
                
                \multirow{2}{*}{Multi-step} 
                & PGD2-AT & \textbf{62.14} & 28.37 & 21.56 & 22.82 & 26.22 & 20.40 & 179.18\\
                & PGD7-AT & 58.88 & \textbf{30.64} & \textbf{25.66} & \textbf{25.62} & \textbf{29.03} & \textbf{23.40} & 467.88\\
                \midrule
                
                \multirow{4}{*}{Multi-stage} 
                & \pname{2} & \textbf{63.82} & 29.55 & 22.59 & 23.64 & 27.31 & 20.34 & 122.90\\
                & \piname{1,2,3} & 63.70 & 29.59 & 23.11 & 24.01 & 27.58 & 20.91 & 124.26\\
                & \pname{5} & 62.64 & 30.45 & 24.28 & 24.82 & 28.64 & 21.69 & 180.88\\
                & \piname{2,5,8} & 61.47 & \textbf{31.42} & \textbf{26.42} & \textbf{26.09} & \textbf{30.07} & \textbf{23.28} & 183.14\\
                \bottomrule
            \end{tabular}
        }
    \end{subtable}
\end{table}

Until recently, Smith~\etal~\cite{smith2019super} discover that cyclic learning rate schedule~\cite{smith2017cyclical} helps the training converge faster. Later on, Wong~\etal~\cite{wong2020fast} adopt cyclic learning into Fast-AT to accelerate the training process and obtain better robustness performance. To further verify the effectiveness of the proposed methods, we carry out the evaluations as in Section~\ref{sec:stepwise} using the cyclic learning rate paradigm. 

On the CIFAR-10 dataset, as reported in Table~\ref{tab:cycle:CIFAR10}, among the single-step adversarial training methods, GradAlign consistently outperforms Kim~\etal against various attacks. With the similar training time cost of Fast-AT, our \piname{1,2,3} exhibits better performance than \pname{2}, which is on par with GradAlign. 
For \piname{2,5,8} which takes the same amount of training cost of PGD2-AT, it outperforms PGD2-AT and GradAlign significantly and achieves comparable performance to PGD7-AT. 

We also summarize the results on CIFAR-100 dataset in Table~\ref{tab:cycle:CIFAR100}, which consistently demonstrate the high effectiveness and efficiency of our proposed methods. Note that Kim~\etal, which aims to avoid catastrophic overfitting by utilizing the minimal scaled adversarial perturbation, cannot prevent catastrophic overfitting on CIFAR-100 dataset. 
We suspect the reason that CIFAR-100 contains more categories but fewer images per class than CIFAR-10, making it more prone to catastrophic overfitting. 
Thus, we argue that CIFAR-100 might be a significant dataset for evaluating catastrophic overfitting in the single-step adversarial training, which is usually ignored in previous works~\cite{andriushchenko2020understanding,vivek2020single,kim2021understanding}. By contrast, our methods do not result in such a phenomenon and \piname{2,5,8} achieves similar robust accuracy with PGD7-AT, which further validates the stability and effectiveness of our methods.




\subsection{Ablation Studies on MOAT}
\label{sec:ablation}

\begin{figure}[tb]
    \centering
    \begin{minipage}[c]{0.48\textwidth}
        \begin{subfigure}{\textwidth}
            \includegraphics[width=\linewidth]{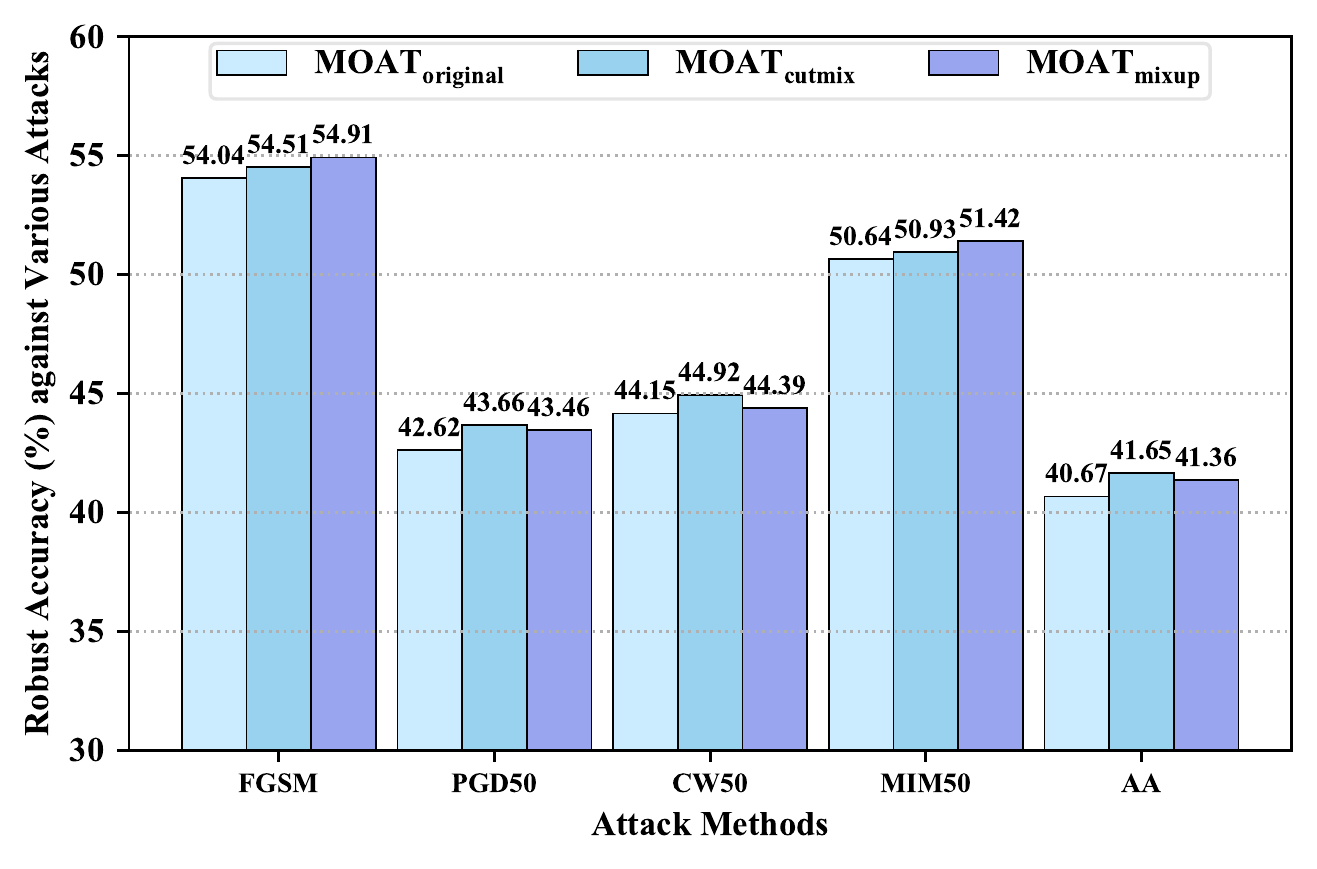}
            \caption{CIFAR-10}
        \end{subfigure}
    \end{minipage}
    \hspace{0.3cm}
    \begin{minipage}[c]{0.48\textwidth}
        \begin{subfigure}{\textwidth}
            \includegraphics[width=\linewidth]{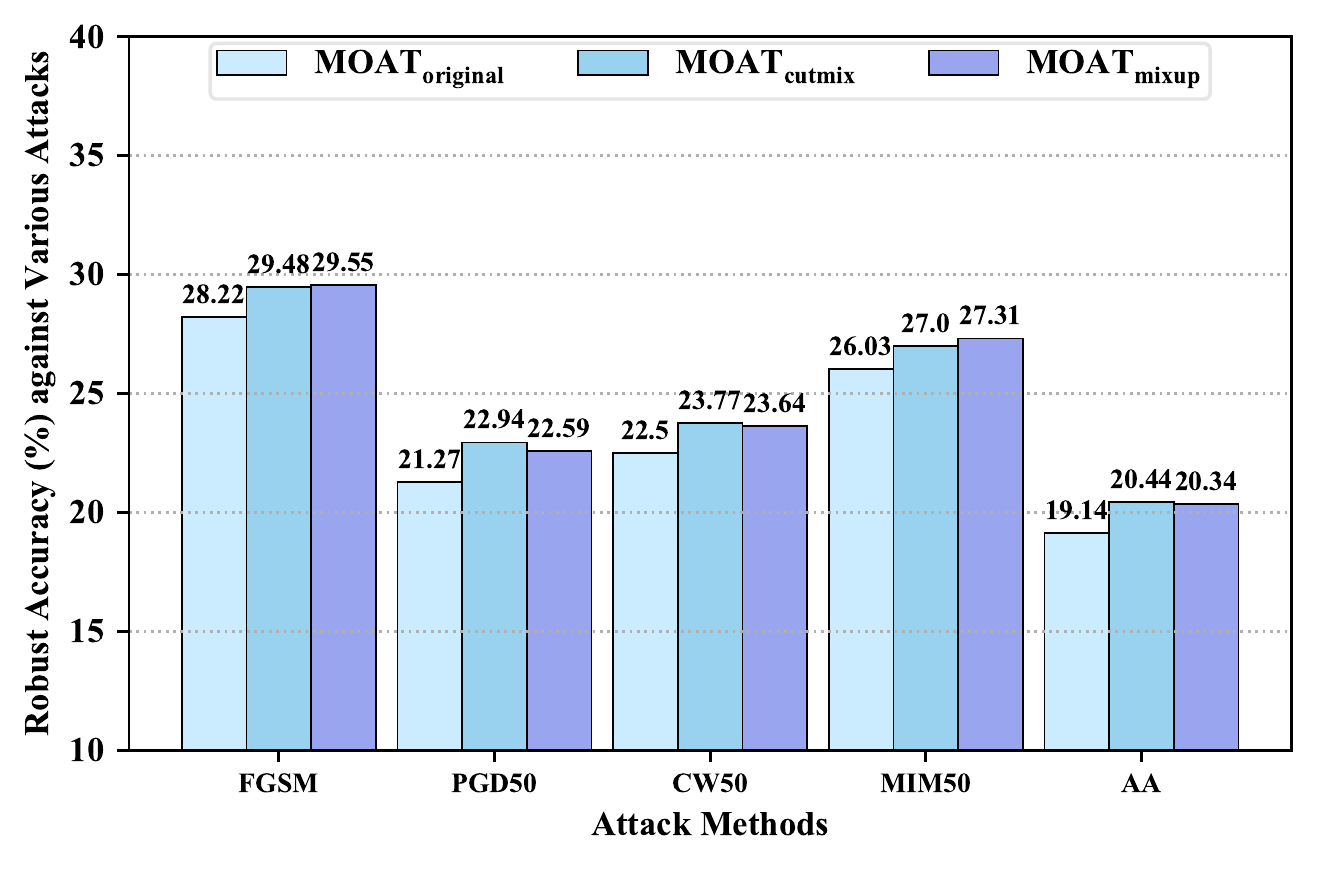}
            \caption{CIFAR-100}
        \end{subfigure}
    \end{minipage}
    \caption{Classification accuracy 
    of MOAT with different samples used at Stage \romman{1} using cyclic learning rate schedule.
    The MOAT methods that adopt original samples, \textit{cutmix} samples, and \textit{mixup} samples at Stage \romman{1} are denoted as \pname{\text{original}}, \pname{\text{cutmix}}, and \pname{\text{mixup}}, respectively. The results under various adversarial attacks indicate that better standard generalization at the first stage of MOAT can help boost the model robustness.
    }
    \label{fig:clean_cutmix_mixup}
\end{figure}

\textbf{The impact of \textit{mixup} at Stage \romman{1}.} As described in Section~\ref{sec:method}, we conjecture that \textit{the better standard generalization could help generate more powerful adversarial examples to improve the adversarial robustness}.
Based on this hypothesis, we adopt \textit{mixup} samples~\cite{zhang2017mixup}, which randomly interpolates two sampled original images and the corresponding labels, at Stage \romman{1} to improve the standard generalization. 
To validate this hypothesis, we adopt original samples and \textit{cutmix}~\cite{yun2019cutmix} samples at Stage \romman{1}, respectively, where \textit{cutmix} is a variation of \textit{mixup} for promoting the standard generalization. 
We denote the MOAT methods that adopt original samples, \textit{cutmix} samples, and \textit{mixup} samples at Stage \romman{1} as \pname{\text{original}}, \pname{\text{cutmix}}, and \pname{\text{mixup}}, respectively.

We evaluate the robustness of these three methods against various attacks on CIFAR-10 and CIFAR-100 datasets.
As shown in Figure~\ref{fig:clean_cutmix_mixup}, on both datasets, \pname{\text{original}} exhibits the lowest robust test accuracy against various attacks.
By contrast, 
when we adopt \textit{cutmix} samples at Stage \romman{1}, \pname{\text{cutmix}} also boosts the model robustness. 
Generally speaking, the performance of \pname{\text{cutmix}} is on par with the one of \pname{\text{mixup}} against various attacks. 
The results provide strong support to our hypothesis, \ie the better standard generalization at Stage \romman{1} can help the subsequent stages find more adversarial examples for training to boost the adversarial robustness.

\begin{wrapfigure}{r}{0.5\textwidth}
    \vspace{-0.55cm}
    \centering
    \begin{minipage}[c]{0.23\textwidth}
        \begin{subfigure}{\textwidth}
            \includegraphics[width=\textwidth]{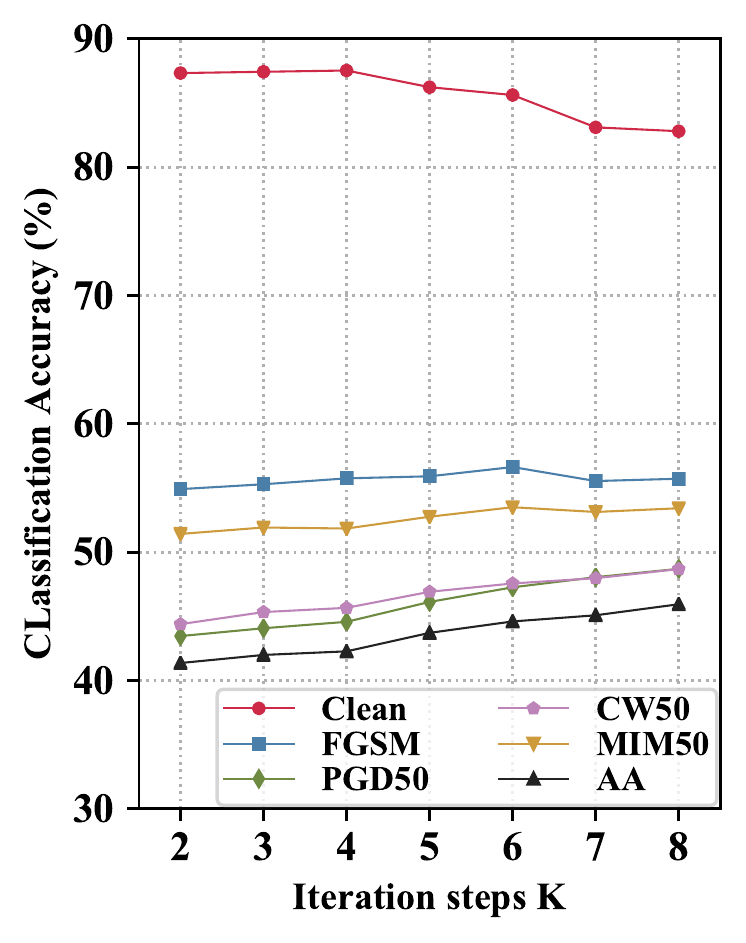}
            \caption{CIFAR-10}
        \end{subfigure}
    \end{minipage}
    \hspace{0.1cm}
    \begin{minipage}[c]{0.23\textwidth}
        \begin{subfigure}{\textwidth}
            \includegraphics[width=\textwidth]{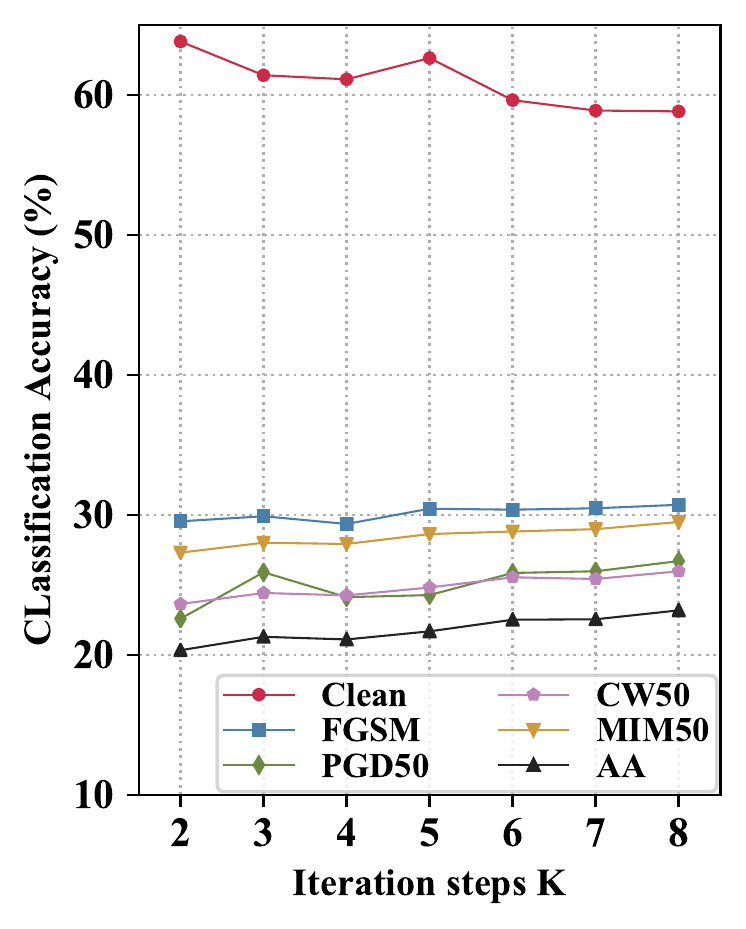}
            \caption{CIFAR-100}
        \end{subfigure}
    \end{minipage}
    \vspace{-0.4em}
    \caption{Classification accuracy (\%) of \name using cyclic learning rate schedule against various adversarial attacks when varying the iteration step $K$ from $2$ to $8$ 
    for the multi-step adversarial examples at Stage \romman{3}.}
    \label{fig:ablation}
    \vspace{-1.3em}
\end{wrapfigure}
\textbf{The Impact of hyper-parameter $K$.} For the proposed method \name, the number of iteration steps $K$ affects the adversarial strength of the multi-step adversarial examples used in Stage \romman{3}. To analyze the impact of the hype-parameter $K$, we vary $K$ from 2 to 8 for the \name method.
As shown in Figure~\ref{fig:ablation}, increasing the value of $K$ leads to better robustness against various attacks but also degrades the standard accuracy.
However, it is noticed that a larger $K$ indicates larger training overhead as discussed in Section~\ref{sec:method}.
In addition, according to the results in Table~\ref{tab:cycle} and Figure~\ref{fig:ablation}, we see that \piname{2,5,8} achieves better performance than \pname{8} with less training overhead, which further validates the high effectiveness of the improved variant \iname.

\section{Conclusion}
In general, catastrophic overfitting is a crucial problem for computationally efficient adversarial training. 
In this paper, we find that adopting multi-step adversarial examples in the training could prevent models from catastrophic overfitting, even for the two-step PGD adversarial examples. 
Motivated by this observation, we propose a novel and efficient adversarial training method called the Multi-stage Optimization based Adversarial Training (\name). 
Unlike most existing adversarial training methods focusing on a fixed type of adversarial examples, \name periodically trains the model on benign examples, single-step adversarial examples and multi-step adversarial examples stage by stage to prevent catastrophic overfitting and reduce the overall training overhead. 
Empirical evaluations on two benchmark datasets, CIFAR-10 and CIFAR-100, validate the high efficiency and effectiveness of the proposed methods on developing adversarially robust models. 
Note that our method is general to other iterative adversarial training methods, and we will incorporate our method into other iterative adversarial training methods in our future work.

\bibliographystyle{plain}
\bibliography{ref}

\newpage
\appendix
\section{The \iname Algorithm}
\label{lab:msoat++}
As we discussed in Section~\ref{sec:imoat}, we consider an improved variant of \name, \iname,  by introducing the increasing step $K$ for the multi-step adversarial examples at Stage \romman{3}. Here we provide the details of \iname in Algorithm~\ref{alg:MSOAT++}.

\begin{algorithm}[h]
    \algnewcommand\algorithmicinput{\textbf{Input:}}
    \algnewcommand\Input{\item[\algorithmicinput]}
    \algnewcommand\algorithmicoutput{\textbf{Output:}}
    \algnewcommand\Output{\item[\algorithmicoutput]}
    \caption{Improved Multi-stage Optimization based Adversarial Training (\iname) }
    \label{alg:MSOAT++}
	\begin{algorithmic}[1]
		\Input $\mathcal{D}$: training samples, $T$: training epochs, $b$: training batch size,
		$\eta$: learning rate,
		$K_1$: number of iteration steps for multi-step adversarial examples at the early training phase,
		$K_2$: number of iteration steps for multi-step adversarial examples at the mid-training phase,
		$K_3$: number of iteration steps for multi-step adversarial examples at the final training phase,
		$\alpha_m$: step size of the multi-step adversarial examples,
		$\alpha_s$: step size of the single-step adversarial examples,
		$\epsilon$: maximum perturbation
		\Output Adversarially robust model $f_\theta$
		\State Randomly Initialize $\theta = \theta_0$
		\For{$t = 0 \ \ {\rm to}\ \  T-1$}
		    \State  Stage $s = t\ \%\ 3 + 1$
		    \For{each mini-batch $(x_{b}, y_{b}) \in \mathcal{D}$}
	        \If{$s == 1$}  \algorithmiccomment{Stage \romman{1}}
	        \State Obtain $(x'_{b}, y'_{b})$ by randomly shuffling $(x_{b}, y_{b})$, and sample $\lambda\sim U(0,1)$
	        \State Mixup $(x_{b}, y_{b})$ with $(x'_{b}, y'_{b})$:
	        \State \qquad $\hat{x}_{b}  = \lambda \cdot x_{b} + (1-\lambda) \cdot x'_{b}$, \quad $\hat{y}_{b} = \lambda \cdot y_{b} + (1-\lambda) \cdot y'_{b}$ 
	        \ElsIf{$s == 2$} \algorithmiccomment{Stage \romman{2}}
	        \State Generate the single-step adversarial training samples:
	        \State \qquad $x_b^0 = \Pi_{\mathcal{B}_\epsilon(x_b)}[x_b + \mathcal{U}(-\epsilon, \epsilon)]$
	        \State \qquad $\hat{x}_b = \Pi_{\mathcal{B}_\epsilon(x_b)}[x_b^0+\alpha_s \cdot {\rm sign}(\nabla_{x_b} \mathcal{L}(f_\theta(x^0_{b}), y_b)]$, \quad $\hat{y}_{b} = y_{b}$
	        \ElsIf{$s == 3$} \algorithmiccomment{Stage \romman{3}}
	        \State Generate the multi-step adversarial training samples:
	        \State \qquad \textbf{if} $t < T/3 $ 
	        \textbf{then} \algorithmiccomment{Increase the step $K$ as training progresses}
	        \State \qquad \qquad $K = K_1$
	        \State \qquad \textbf{else if} $t < 2T/3 $
	        \textbf{then}
	        \State \qquad \qquad $K = K_2$
	        \State \qquad \textbf{else if} $t < T $
	        \textbf{then}
	        \State \qquad \qquad $K = K_3$
	        \State \qquad $x_b^0 = \Pi_{\mathcal{B}_\epsilon(x_b)}[x_b + \mathcal{U}(-\epsilon, \epsilon)]$, \quad $\alpha = \max(\alpha_m, \epsilon/K) $
	        \State \qquad \textbf{for}\ \ $k = 1$ \ \ {\rm to}\ \ $K$  \textbf{do}
	        \State \qquad\qquad $x^k_{b}=\Pi_{\mathcal{B}_\epsilon(x_b)}[x_b^{k-1}+\alpha \cdot {\rm sign}(\nabla_{x_b} \mathcal{L}(f_\theta(x^{k-1}_{b}), y_b)]$
            \State \qquad \textbf{end for}
	        \State \qquad $\hat{x}_b = x^K_b$, \quad $\hat{y}_{b} = y_{b}$
	        \EndIf
	        \State $\theta_{t+1} \leftarrow \theta_{t} - \eta \cdot  \nabla_{\theta_t} \mathcal{L}(f_{\theta_t}(\hat{x}_{b}), \hat{y}_{b})$  \algorithmiccomment{Update the model parameters}
	        \EndFor
		\EndFor
	
	\end{algorithmic} 
\end{algorithm} 

\end{document}